# Interdisciplinary Workshop on Mechanical Intelligence: Summary Report

May 30th, 2024 to May 31st, 2024
Held at the National Science Foundation
2415 Eisenhower Avenue, Alexandria, VA 22314

## List of Participants and Report Authors

| Participants | Affiliation |
| --- | --- |
| Aaron Johnson | Carnegie Mellon University |
| Aja Mia Carter | Carnegie Mellon University |
| Amir Alavi | University of Pittsburgh |
| Andres F Arrieta | Purdue University |
| Anthony Bloch | University of Michigan |
| Barry Trimmer | Tufts University |
| Brian Do | Oregon State University |
| C. Chase Cao | Case Western Reserve University |
| Chen Li | John's Hopkins University |
| Emma Lejeune | Boston University |
| Eva Kanso | University of Southern California |
| Hannah Stuart | University of California at Berkeley |
| Jean-Michel Mongeau | Penn State University |
| Jeffrey Lipton | Northeastern University |
| Jordan Raney | University of Pennsylvania |
| Kaushik Jayaram | University of Colorado Boulder |
| Kon-Well Wang | University of Michigan |
| Laura Blumenschein | Purdue University |
| Mahdi Haghshenas-Jaryani | New Mexico State University |
| Margaret Coad | University of Notre Dame |
| Mark Cutkosky | Stanford University |

| Mark Yim | University of Pennsylvania |
| Matt McHenry | University of California, Irvine |
| Michael Dickey | NC State University |
| Michael Tolley | UC San Diego |
| Nick Gravish* | UC San Diego |
| Paolo Celli | Stony Brook University |
| Rob MacCurdy | University of Colorado, Boulder |
| Bob Full | University of California at Berkeley |
| Ryan D. Sochol | University of Maryland, College Park |
| Sarah Bergbreiter | Carnegie Mellon University |
| Suyi Li | Virginia Tech |
| Todd Murphey | Northwestern University |
| Tony Chen | Stanford University |
| Victoria Webster-Wood* | Carnegie Mellon University |
| Wenlong Zhang | Arizona State University |
| Wenzhong Yan | UCLA |
| Zeynep Temel | Carnegie Mellon University |
| T.J. Wallin | MIT |
| Mark Plecnik | University of Notre Dame |

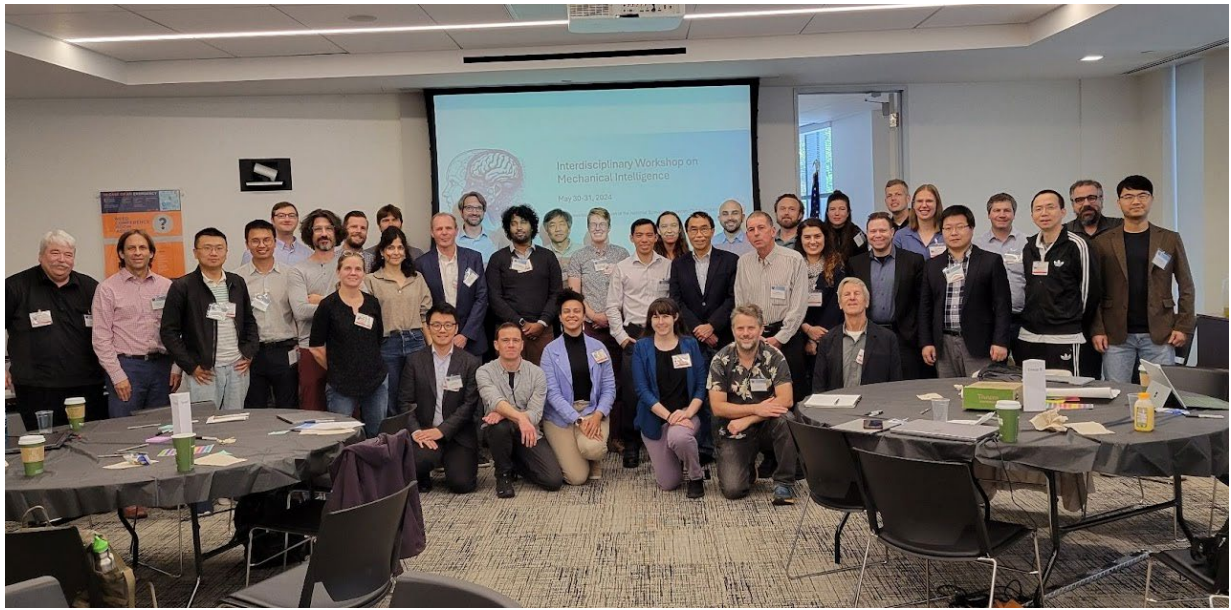

# Executive Summary

This report provides a summary of the outcomes of the Interdisciplinary Workshop on Mechanical Intelligence. Mechanical Intelligence (MI) represents the phenomenon that novel structural features of material/biological/robotic systems can encode intelligence through responsiveness, adaptivity, memory, and learning in the mechanical structure itself. This is in contrast to computational intelligence wherein the intelligence functions occur through electrical signaling and computer code. The two-day workshop was held at NSF headquarters on May 30-31 and included 38 invited academic researcher participants, and 8 program officers from the NSF. The workshop was structured around active small and large group discussions in groups of 4-5 and 9-10 with the goal of addressing topical questions on MI. Working groups entered notes into shared presentation slides for each discussion session and presented their outcomes in a final presentation on the last day. To motivate/stimulate discussions four keynote speakers presented 25 minute presentations that addressed questions on MI and incorporated their work and perspectives into MI relevance.

Here we summarize the overall outcomes of the workshop. This workshop was focused on developing consensus definitions of mechanical intelligence and how it can be quantified, tested, compared to algorithmic intelligence, and what are the possible future opportunities for MI. In the discussions from the first keynotes and breakout sessions there did not appear to be a clear consensus on "what is mechanical intelligence". A debate about how adaptive a system must be to display MI took place after the first keynotes with the two positions diverging on whether a system like a mechanical cam or watch is mechanically intelligent (as introduced by Barry Trimmer's keynote). This led in well to the breakout session discussion in which we asked teams to delineate between pure mechanics, mechanical intelligence, and pure algorithmic intelligence. Groups had a hard time distinguishing between these regimes and healthy debate ensued. However, despite these disagreements, there was broad consensus that mechanical systems can display intelligent behavior. Without diving into a debate about what constitutes intelligence, the workshop attendees broadly characterized intelligence as systems that display robustness, adaptivity, memory, and learning capabilities and specifically mechanically intelligent systems do this through means of structural systems and their kinematics/dynamics/physics of movement and environmental interaction.

The lack of singular consensus on mechanical intelligence led to the idea that mechanical intelligence exists on a spectrum of possible behaviors and complexity. Furthermore, it was suggested that mechanical intelligence need not be a noun, but instead is a property of a system as an adjective. For example mechanically intelligent robot hands, mechanically intelligent sensors, mechanically intelligent metamaterials, etc. This recasting of mechanical intelligence as a feature of a system better captures the wide breadth and scope of MI research and doesn't pigeonhole mechanical intelligence to a single domain like robotics or metamaterials for example. The concept of mechanical intelligence on a spectrum also demands the identification and adoption of consistent metrics or benchmarks in order to compare different mechanically intelligent systems. However, the participants agreed that there is no singular axis upon which to measure mechanical intelligence, and instead the field needs to consider reporting metrics across multiple dimensions or axes.

Participants were overall enthusiastic about the future possibilities of mechanical intelligence in future research and application. It was stated that barriers to future progress were largely focused on funding, standardization, lack of integrated design tools, lack of an overall theory or framework, manufacturing limitations, and the challenges of initiating and sustaining large interdisciplinary collaborative efforts. There were many encouraging and interesting proposed applications of mechanical intelligence discussed by the teams including (but not limited to) low- or no-power mechanical subsystems, unhackable robotic systems, sustainable and biodegradable intelligent systems, and mechanical systems that adapt behavior to their environmental context. One of the biggest ideas to emerge from this workshop is that mechanical intelligence does not need to be considered separate from computational intelligence, but instead, one of the biggest areas of opportunity is how to best integrate computational intelligence (i.e. algorithmic control) with mechanically intelligent subsystems. The enthusiasm for this idea may best be illustrated by Group A's final report in which they generated a plot (see below) of mechanical/computational intelligence for legged robotics. They highlighted the specific technology gap as existing between these two extremes. Participants thought this was very exciting and next noted that by taking a complementary (rather than adversarial) design approach with computational machine intelligence there would be many future opportunities for hybrid mechanically and electrically intelligent systems.

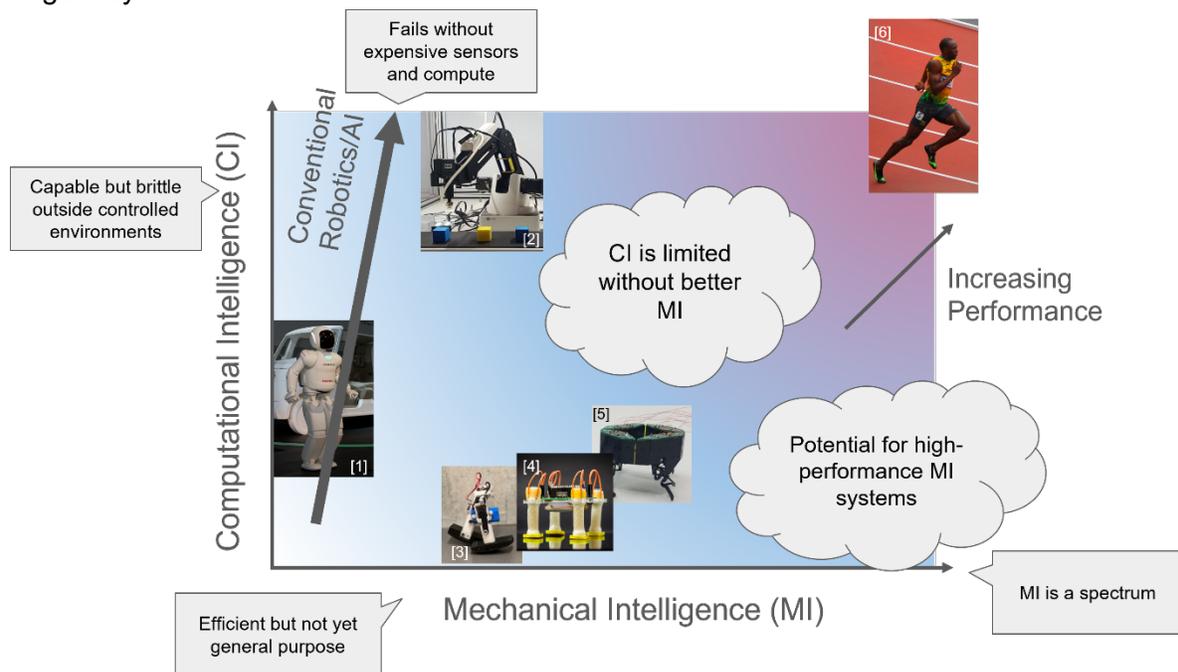

*Figure 1. Participant generated visual representation of the continuum of mechanical and computational intelligence. [1] Tokyo Motor Show 2011: ASIMO (version 2011) photo by Morio CC-BY-SA 3.0. [2] Dobot- A robot arm used for applications like color sorting, stacking, laser printing, etc. photo by DangerAlpha CC0. Photo of "Zippy" courtesy of Dr. Aaron Johnson. [4] "UCSD-JacobsSchool-20191001-Shengqiang_walking_robot-01650-8MP" by UC San Diego Jacobs School of Engineering, CC BY 2.0. [5] Kabutz, H. and Jayaram, K. (2023), Design of CLARI: A Miniature Modular Origami Passive Shape-Morphing Robot. Adv. Intell. Syst., 5: 2300181. https://doi.org/10.1002/aisy.202300181. CC-BY 4.0. [6] "Run" by James Mackintosh Photography is licensed under CC BY 2.0.*

# Technical Report Generation

Throughout the workshop, participants took notes in Google Slides and paper easels. Presentations were recorded on Zoom for transcription of the presentation and discussion using the built-in transcription capabilities. All notes were organized by topic area, and prompt and consensus identification was performed using Microsoft Copilot. Generated consensus content was reviewed and edited by the workshop organizers and report authors to ensure that the notes had been processed properly and accurately reflected the workshop discussion.

# Schedule of workshop activities

**May 30th**
| | |
|---|---|
| 8:00am | Breakfast (provided) |
| 8:30-9:00 | Welcome, Vision and Scope |
| | Drs. Nick Gravish and Vickie Webster-Wood |
| | NSF Opening Remarks |
| 9:00-10:00 | Keynotes |
| | Barry Trimmer, Tufts University |
| | Hannah Stuart, UC. Berkeley |
| 10:00-10:30 | Discussion and challenge questions |
| 10:30-11:00 | Break |
| 11:00-12:30 | Small Group Discussions |
| 12:30-2:00 | Lunch (provided) |
| 2:00-3:00 | Keynotes |
| | Sarah Bergbreiter, Carnegie Mellon University |
| | Todd Murphey, Northwestern University |
| 3:00-3:30 | Discussion and challenge questions |
| 3:30-5:00 | Small Group Discussions |
| | |
| 6:30-9:00 | Dinner at Hummingbird Bar and Kitchen |
| | 220 S. Union Street, Alexandria, Virginia, 22314 (Inside Hotel Indigo) |

**May 31st**
| | |
|---|---|
| 8:00am | Breakfast (provided) |
| 9:00-12:30 | Small Group Discussions |
| 12:30-1:30 | Lunch (provided) |
| 1:30-3:00 | Report Outs |
| 3:00-3:30 | Closing remarks |

## List of NSF attendees

| | |
|---|---|
| Jordan Berg | Program Director, CMMI FRR |
| Alex Medina-Borja | Program Director, CMMI Human-Centered Engineered Systems |
| Floh Thiels | Program Director, IOS |
| Dan Linzell | Division Director, Civil, Mechanical and Manufacturing Innovation Division, Directorate for Engineering |
| Sohi Rastegar | Head, Office of Emerging Frontiers and Interdisciplinary Activities, Directorate for Engineering |
| Michelle Elekonich | Deputy Division Director, Division of Integrative Organismal Systems, Directorate of Biological Sciences |
| Siddiq Qidwai | Acting Deputy Division Director CCF |
| Alexander Leonessa | Program Director, CMMI FRR M3X |

# Summary of break out discussion session 1

The first break-out session involved a series of activities to help the participants focus in on the bounds and definitions of mechanical intelligence.

## 1. Activity 1: Trying to identify the relationships between common terms used in mechanical intelligence discussions

Activity 1 asked each group to attempt to discuss and identify the relationship between embodied intelligence, physical intelligence, mechanical intelligence, and morphological computation. The responses to this activity were highly varied across groups. The table below shows the responses from each group.

| Group 1 | Group 2 |
|---|---|
| Embodied Intelligence, Mechanical Intelligence, Morphological Computation, Physical Intelligence (shown as scattered/overlapping boxes) | Embodied Intelligence, Physical Intelligence, Mechanical Intelligence, Morphological Computation (listed vertically)<br><br>Are these terms interchangeable?<br><br>Do we have anything that qualifies as intelligence? Intelligence vs computation<br><br>Physical forces: electro-magnetic, chemical, contact, …<br><br>Is "morphology" just geometry or also materials |

| Group 3 | Group 4 |
|---|---|
| **A hierarchy of terms**<br>Hierarchy of intelligence (degrees of responsiveness scale)<br><br>Central Nervous system to Peripheral motor control<br>Centralized (thoughtful/varied) ←→ Low-level (reflexiv/predictable)<br>Very intelligent (goal changes)  Memory  Simple logic  Responsive material  A rock<br><br>Features:<br>Sensing  Seasonal adaptation  Thermostat  Thermometer<br>Adaptation over time  A spring<br>Computing  Neural network  Abacus<br><br>Embodied Intelligence — Mechanical Intelligence — Physical Intelligence  Computation: – CPU, GPU, Mechanical (Morphological) Computing<br><br>"Intelligence" is a branding method to contrast with AI. | Embodied Intelligence containing Mechanical Intelligence, Physical Intelligence, and Morphological Computation (as overlapping nested boxes) |

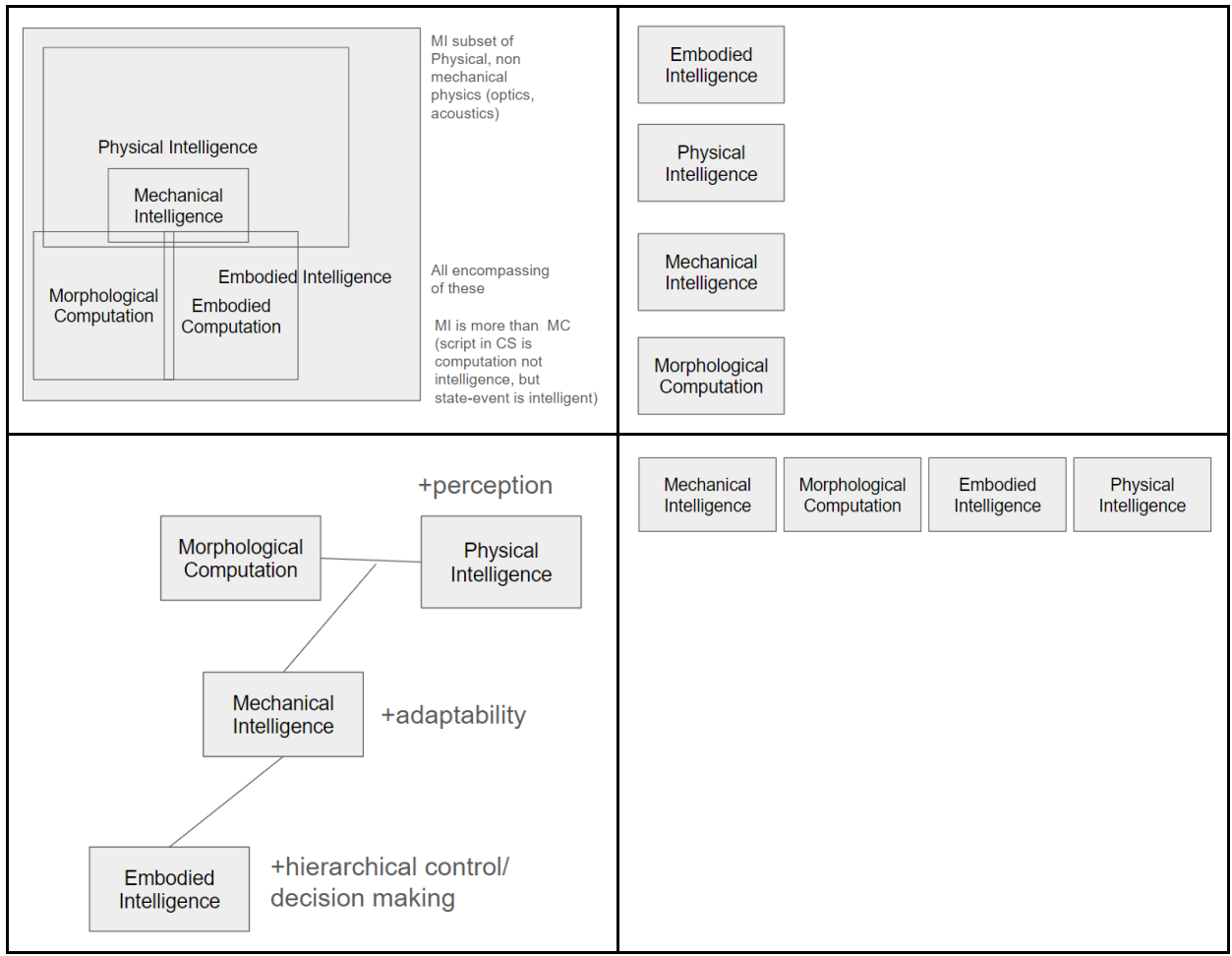

## 2. Identify examples that define the bounds of mechanical intelligence

Activity 2 asked each group to provide examples that fall along a spectrum where the left side is pure mechanics, the middle is mechanical intelligence, and the right side is pure algorithm. Based on clustering of these lists, the following definitions and examples were extracted:

**Pure Mechanics:** These are systems that perform the same action every time, given the same situation, even when it fails. They are like the "transducers" or "building blocks" that alone are not intelligent, or only intelligent at the lowest level. Examples include clocks, figurines, automaton with no memory, passive vibration isolation mount, pulley systems, piezoelectric transducer, mechanical components, wave propagation, structural properties, tuned frictional properties, inertial properties, fracture, linkages/cams, gear system, internal combustion engine (thermal cycle), mechanisms, Newton's laws, a mechanical switch, rigid bodies, pendulum, simple machines, and a prescribed/forced motion.

**Between Pure Mechanics and Mechanical Intelligence:** These are systems that exhibit some level of mechanical intelligence but are not purely algorithmic. Examples include passive dynamic walkers, flyball governor, and Venus Fly trap.

**Mechanical Intelligence:** These are systems that can do different things in response to the same inputs, based on learning/adaptation. They exhibit emergence, adaptability, responsiveness, robustness, and decision-making capabilities. Examples include underactuated gripper, metamaterial adaptable wave guides, shape memory, hysteresis, bistable/multistable structures, reservoir computing, passive conforming hands/squirrel feet, passive dynamic walker, underactuated hands, intentionally harnessed geometric instabilities, mechanical metamaterials, programmed instabilities, physical reservoir computing, flyball governor, Rhex legs climbing, thermostat, underactuated grippers, and watched clock (where intelligence comes from the use i.e., clock plus human is intelligent).

**Pure Computational Algorithm:** These are systems that have no ability to act on the world and no embodiment, i.e., anything that can achieve these, but that doesn't have a mechanical component. Examples include digital computation, computer code, trajectory optimization, LLMc, data sorting, computer vision, matrix multiplication, numerical integration, goal setting, Chat GPT, calculator, off-robot/system control schemes, neural network, reservoir computing, and ChatGPT (which is considered really intelligent).

# 3. Identify how mechanical intelligence can be quantified in experiments, simulation, and theory

Activity 3 asked participants to identify quantitative tests for assessing mechanical intelligence in (1) experiments, (2) simulations, and (3) theory. Participants consistently noted that reproducibility and standardization were needed but challenging.

## Experiments:

The participants suggested several ways to quantify Mechanical Intelligence (MI) in experiments:

1. **Perturbation:** Introduce changes to the system and observe its response.
2. **Contextual Appropriateness:** The system's response should be appropriate for the specific task at hand.
3. **Diversity:** Measure the range of different outcomes or tasks the system can handle.
4. **Robustness:** The system should be able to handle uncertainty and continue to perform well.
5. **Mechanical Turing Test:** Similar to the Turing test in AI, this would assess if a mechanical system's intelligence is indistinguishable from human intelligence.
6. **Efficiency:** This could be measured in terms of energy use, computational power, or the number of lines of code needed to accomplish a task.
7. **Speed:** How quickly the system can complete a task.
8. **Resilience:** The system's ability to recover from or adjust easily to changes.

9. **Relation between Inputs and Outputs:** For example, in grasping, inputs could be contact conditions, shape of the object, total energy, and outputs could be error, accuracy, speed, force, adaptability, efficiency, and computation power.
10. **Ethical Impacts and Sustainability:** Consider the broader impacts of the system, including ethical implications and sustainability.

It was also noted in the discussion that simulations are numerical experiments, and robustness is not about optimization but sufficiency. The design and implementation of accessible benchmark tasks with strong baselines is a major challenge that is time-consuming and hard to fund. The most useful measurements are those that apply to real environments.

## Simulation:

For the groups that did not note that simulations were the same as experiments:

1. **Emergence:** Measure the emergence of complex behavior or properties from simple rules.
2. **Controlled Randomness:** Introduce randomness in a controlled manner to test the system's response.
3. **Digital Twins:** Create a digital replica of a physical system to analyze its performance.
4. **Virtual Experiments of Task Success:** Conduct virtual experiments to measure the success rate of tasks.
5. **Out of Distribution with a Good Enough Model:** Test the system with data that's not part of the training set to verify its robustness.
6. **Measure the Impact of Arbitrary Material Properties:** The ability to input arbitrary material properties without having to realize them physically
7. **Virtual Environment AR/VR:** Use augmented reality or virtual reality environments for simulations
8. **Robustness and Bounds of System:** Using simulations to determine the robustness and bounds of the system.

However, participants also expressed concerns regarding the limits and inaccuracies of simulations.

## Theory:

The participants suggested several theoretical approaches to quantify Mechanical Intelligence (MI):

1. **Computational Complexity:** Something akin to computational complexity could be used to measure MI.
2. **Number of Internal States/Memory:** The number of internal states or memory could be a measure of MI.
3. **Statistical Mechanics and Perturbation Theory:** These theories could be used to model and quantify MI.

4. **Neuroscience and Control Theory:** Insights from these fields could be applied to MI.
5. **Plasticity and Evolutionary Theory:** These concepts could be relevant in the context of adaptable mechanical systems.
6. **Information Theory:** This could provide a framework for quantifying MI, including concepts like entropy.
7. **Machine Learning Theory:** This could be applied to MI, for example, how entropy can inform machine learning decision making.
8. **Generalizability:** The ability of a system to perform well on unseen data.
9. **Task-Based Comparisons:** Comparisons to non-MI baselines on specific tasks.
10. **Analogies to Algorithmic Computing:** Drawing parallels between MI and traditional computing could provide insights.
11. **Lyapunov Stability:** Measures like Lyapunov stability could be used to put guarantees on the outputs over a range of inputs.
12. **Mathematical and Cognitive Models:** These could provide theoretical frameworks for understanding and quantifying MI.
13. **Measurement of MI IQ:** The concept of an IQ for MI was suggested.
14. **Energy Transduction:** How energy moves from mechanical to intelligence could be a measure of MI.
15. **Objective Function:** An objective function could be defined to quantify the level of intelligence, though it was noted that this is difficult and different types of intelligence may require different functions.
16. **Change of Intelligence:** The focus could be on a change of intelligence ("delta" intelligence or sensitivity studies), not absolute intelligence, which is highly contextual.

The participants also discussed the definition of intelligence, asking whether it is analytical, practical, or creative, and how many decisions are being made. They noted that these factors could influence the quantification of MI.

## 4. Discussion Session 1 Takeaways

The key takeaways from the discussion on Mechanical Intelligence (MI) during Session 1 are as follows:

- **Continuum of MI:** A continuum of MI was proposed, ranging from sensing/actuation to memory and learning. This led to the idea of Mechanical Intelligence not as a noun, but rather as an adjective (e.g. mechanically intelligent sensors).
- **Dimensions and Levels of MI:** The dimensions of MI include space, time, environment, task, hierarchy, etc. There are levels of mechanical intelligence, with robustness, memory, and computation and learning on that memory being key components.
- **Benchmarks for MI:** The need for benchmarks to quantify MI was highlighted. Quantifying mechanical intelligence is important, but benchmarks are non-standardized.
    - **Task Performance vs Robustness:** These were considered as potential metrics for MI.

- ○ **Standards for Quantifying MI:** The need for standards (e.g., standardized experiments adopted by the community) for quantifying MI was emphasized. Defining useful benchmarks is challenging and will require money, expertise, and time. Currently, where there are not clear benchmarks, researchers should place extra emphasis on reproducibility of their results.
- **Intelligence and Computation:** Intelligence and computation are not synonymous, and the term "intelligence" is currently controversial and poorly defined across multiple disciplines.
- **Mechanics and Complex Behavior:** Mechanics alone can explain highly complex behavior, and the transition to "mechanical intelligence" will come with intentionally harvesting and programming these non-linear responses.
- **Biology and Bio-inspiration:** Biology and bio-inspiration were considered important by participants in the context of MI.
- **What is Intelligence?:** The definition of intelligence, including aspects like memory, learning, and the ability to exhibit different behaviors in response to the same stimulus, was discussed.
- **Adaptation vs Intelligence:** The participants raised the question of what is adaptation vs. intelligence but did not reach a consensus.

The participants also highlighted the importance of accessible complex designs via 3D Printing for complex behavior and the need for funding for MI research, especially for systems that interact with the "real world". They also discussed the potential benefits of using the term computation rather than intelligence.

# Summary of break out discussion session 2

The second breakout session asked participants to consider metrics and benchmarks that could be used to assess mechanically intelligent systems. They were also asked to think about the insight or impact of such metrics and the barriers that are currently preventing researchers in the field from implementing benchmarks. We have summarized their responses here.

## 1. What metrics can be used to quantify mechanical intelligence

Participants proposed a wide array of performance metrics that could be used to quantify and compare mechanically intelligent systems.

**Performance Metrics:**

- Task Performance (including speed, accuracy, precision, and efficiency)
- Robustness (including resilience to damage and variations in inputs)
- Durability
- Stability
- Maneuverability
- Adaptability (including ability to handle unforeseen scenarios and rapidly adapt in new environments)
- Energy and Computational Efficiency
- Data Efficiency

- Quality and Type of Observations Needed to Perform the Task
- Actuator Characteristics (number, type, and precision of actuation)
- Sensor Characteristics (number and type)
- Collocation of Sensors and Actuators
- Probabilistic Description of Success/Failure/Performance Metrics
- Multi-tasking and Hierarchical Tasking
- Encoding Adaptation/Learning/Decision-Making on a Structure
- Reduced-Order Representation of Mechanical Information
- Interface with Higher/Lower Level of Control/Learning/Cognition
- Environment Handling (extreme/unusual conditions)
- Morphability (changes in size/shape)
- Cost (including manufacturing cost)
- Manufacturability
- Reliability of Outcomes
- Control Authority (max and average)
- SWaP-C: Size, Weight, Power, Cost
- Response Time
- Bandwidth Required
- Error Handling
- Mechanical Energy
- Time Delay
- Safety/Failure Factors (stress/strain)
- Controllability and Observability
- Points of Connection/Points of Failure
- Modularity vs Integrated Design
- Reprogrammability
- Designability
- Safety Measures
- Sustainability/Life Cycle
- Learning Speed and Complexity
- Unlearning/Forgetting
- Cross-Domain Learning
- Complexity (as in Information Theory)
- Total Number of Commands Required
- Long Term Error of the Response
- Stability Margin
- Open Loop Control
- Scalability
- Variance of Input vs. Output
- Human Factors
- Repeatability
- Number of Simulation Cycles
- Number of Bits Neurons
- Eigenvalues (linearize dynamics)
- Objective Performance
- Multiple Objectives
- Reaction Time
- Physical Resilience
- Entropy
- Damage Rate
- Physical Mass of Computer Hardware/Robot Mass
- Energy Output
- Energy Harvesting Efficiency
- Observability of Functional Objective
- Money Cost

2. What are the specific insights or impacts of these various metrics?

While not all groups responded to this prompt, a summarized list of those who did is included here:

- **Error**: Helps in predicting outcomes, validation, and verification.
- **Mechanical Energy**: Takes into account the cost of transport.
- **Time Delay**: Impacts closed-loop stability.
- **Stability**: Ensures reliable response with uncertainties.
- **Safety/Failure Factors**: Helps to avoid breaking rigid parts.
- **Controllability and Observability**: Enables analysis of the system.
- **Traditional Measures of Performance** (such as efficiency, speed, precision, cost, etc.): Used to evaluate performance and trade-offs.
- **Manufacturability**: Determines if we can make what we want and what new manufacturing tools we need.
- **Points of Connection/Points of Failure**: Affects "Robustness" and durability.
- **Modularity vs Integrated**: Impacts ease of debugging and replaceable components.
- **Reprogrammability**: Enables learning and adaptability.
- **Designability**: MI systems often take advantage of nonlinearities, contact, etc. that are hard to model/design. This impacts the ability to model/design/predict behavior and allows for mechanical intelligence for all.
- **Safety**: Determines various applications that it can be applied to (especially human-centric) and security (e.g., cybersecurity).
- **Lines of Code**: Indicates software complexity.
- **Sustainability/Life Cycle**: Affects environmental impact and reuse.
- **Number of Simulation Cycles**: Impacts learning efficiency and effectiveness of learning.
- **Number of Bits Neurons**: Indicates information content.
- **Eigenvalues, Linearize Dynamics**: Affects stability and convergence behavior.
- **Objective Performance**: Affects robustness to perturbations.
- **Multiple Objectives**: Determines the ability to generalize.
- **Reaction Time**: Affects the speed of response.
- **Energetic Efficiency**: Indicates efficiency.
- **Physical Resilience**: Determines the ability to return to the initial state.
- **Entropy**: Indicates information content in transduction, transmission, and memory.
- **Damage Rate**: Affects safety.
- **Physical Mass of Computer Hardware/Robot Mass**: Affects mechanical/computational efficiency.
- **Observability of Functional Objective**: Determines the ability to detect the state.
- **Money Cost**: Indicates efficiency of finances.

## 3. Many control and AI disciplines have the concept of benchmark tests or datasets. What would be the equivalent for mechanical intelligence?

Participants identified and converged on several categories of benchmarks that could be used to assess and compare mechanical intelligence in a given system. Many groups noted the importance of sharing these benchmarks to allow groups to run comparisons more easily with new systems.

The participants at the workshop identified several types of benchmarks for assessing Mechanical Intelligence (MI). Here's a consolidated summary of their discussion:

**Task-Specific Benchmarks:**

- These benchmarks involve problem/task-specific benchmarks, responsiveness to various standard stimuli, and task difficulty.
- These benchmarks test the system's performance on specific tasks such as for a robot: perching, pick & place tasks, walking over rough terrain, memorizing or encoding mechanical information about the environment, spreading peanut butter and jelly on bread, and solving mazes. They also include specifying resource-constrained systems such as delivering a small (micron-scale) package deep into an organ without damaging it.

**Environment-Specific Benchmarks:**

- These benchmarks involve testing MI in variable terrains (rocky, sandy, muddy), and established ranges of terrain to cross. For example, a benchmark may require locomotion over a standard set of terrains.

**Future-Oriented Benchmarks:**

- These benchmarks involve setting future goals, such as achieving certain capabilities in manipulation, locomotion, and negotiating complex mechanical environments in a given timeframe.

**Comparative Benchmarks:**

- These benchmarks involve comparing different MI systems by holding some factors constant while varying others, or comparing to the performance levels of animals and humans.

**Robustness, Complexity, and Efficiency Benchmarks:**

- These benchmarks test the system's robustness to mechanical perturbations, total system complexity, data dimensionality, and energy expenditure. They also include benchmarks related to the size and weight of the system, the amount of work that can be done with a standard amount of energy, and the motivation for reduced computation.
- This category of benchmark should also involve measuring repeatability, possibly combined with perturbation tests.

**Controller Modifiability:**

- These benchmarks involve systems that are not purely mechanical, where the modifiability of the controller and how the mechanical system affects performance are considered important.

**Interaction Benchmarks:**

- These benchmarks involve interactions with set surfaces/materials of varying stiffness, and mechanical signal propagation in heterogeneous materials under both static and dynamic conditions.

**Cognitive Measures:**

- These benchmarks involve defining an IQ for mechanical intelligence, including standard cognitive measures like problem-solving ability, reasoning, adaptation, sensing, perception, etc.

**Competition-Based Benchmarks:**

- These benchmarks involve a "DARPA Grand Challenge" type of competition with a computational limit to a mechanical challenge.

## 4. What barriers prevent you from implementing the benchmarking on the prior slide? These may be institutional, resources, or existing research gaps in a given field (etc.)

As expected, the participants all strongly voiced support for established metrics, and yet to date, the establishment and use of such metrics has been sparse. Workshop participants identified the following barriers that prevent the broad use of benchmarks currently.

**Research Gaps:**

- Lack of baseline physical embodiments of mechanical intelligence for comparison.
- Difficulty in creating "generalizable" benchmarks that allow comparisons of systems without prescribing a particular design/morphology/strategy.
- Difficulty in replicating a state-of-the-art device for a slightly different task due to high costs and complexity in mapping across tasks.
- The need to generate a dataset per device and task, which can be challenging.

**Interdisciplinary Challenges:**

- The need for interdisciplinary team building and a framework for facilitating discussion between dissimilar fields involved in MI, such as information theory, manufacturing, structures, etc.

- Difficulty in aligning many people on a task with metrics.

**Task Specificity:**

- All benchmarks are task-specific, and devices are made for a task. The set of possible tasks is limitless, and the metric set for any task is specific. There may not be total overlap between devices for different tasks.

**Institutional Barriers:**

- The overall incentive structure of academia doesn't inherently reward the pursuit of collective knowledge, rather there are incentives (high impact papers, funding cycles) that encourage siloed approaches.
- Difficulty in standardizing benchmarks among a broad diversity of systems.
- Difficulty in developing a benchmark that is so compelling that others adopt it willingly. This is a major challenge in and of itself and should be appreciated as a research question in its own right.

**Technical Challenges:**

- Harder to compare physical systems than algorithms.
- Soft robots are not standardized like robotic arms. Manufacture is bespoke and under-reported even for 3D printed robots.
- The need for testing as a service.
- Difficulty in defining intelligence.
- The need for several benchmarks to get at the heart of performance.

These barriers provide a comprehensive understanding of the challenges faced in evaluating the performance and capabilities of mechanically intelligent systems. They cover a wide range of aspects, from research gaps and interdisciplinary challenges to task specificity and institutional barriers, among others. This allows for a thorough and holistic understanding of the challenges in this field.

# 5. Discussion Session 3 Takeaway Summary

Potential metrics and benchmarks that could be applied to mechanically intelligent systems:

1. **Control Metrics**: Metrics that are the same as controls, reflecting the system's ability to manage its own operations.
2. **Design Metrics**: Metrics that are the same as design, assessing the system's structural and functional aspects.
3. **Interaction Metrics**: Metrics that go beyond control and design to look at the system's interaction with its environment.
4. **Cognition Level**: Benchmarking mechanical properties while also considering the level of cognition, or the system's ability to acquire and process information.

5. **Data Reduction**: The degree to which the system can reduce the size of the dataset or the power required to synthesize the data.
6. **Variance of Input to Output**: A general metric for comparing systems, measuring how much the output varies in response to changes in the input.
7. **Learning Efficiency**: The ability of the system to learn from a diversity of experiences, and the ability to un-learn over time.
8. **Creativity**: A subjective metric that could be assessed through user surveys or polls, reflecting the system's ability to generate novel and valuable outputs.
9. **Performance Metrics**: Metrics that are easy to devise and reflect the system's ability to perform its tasks effectively and efficiently.
10. **Life Cycle and Sustainability**: Metrics that assess the system's performance over its entire life cycle and its adherence to sustainability principles.

# Summary of break out discussion session 3

In the third breakout session, we asked expert participants to address 4 prompts. We have summarized their responses here by prompt and theme.

## 1. Where is mechanical intelligence going to take us in 5, 10, or 20 years?

The key themes of the future of Mechanical Intelligence (MI) can be summarized as follows:

**1. Advanced Robotics:**

- Robots that interact robustly with messy environments and have high dexterity in multiple unstructured environments.
- Development of general-purpose machines and advanced special-purpose machines.
- Robots that are more agile and able to operate in challenging environments and perform complex tasks.
- Swarm of robots with purely mechanical intelligence.

**2. Bio-Inspired Design:**

- Organismal Robotics, Synthetic Neuromechanics, and Integrative Mechano-biology.
- Biohybrid/biointegrated robots that are sustainable and biocompatible.
- Mechanically learned/adapted personalization of morphology for household tasks/wearables.

**3. Medical Applications:**

- Surgical tools that can operate in complex working conditions and challenging constraints.
- Medical robots that can work in small scale/constrained environments (inside human body, MRI).

- MRI-Compatible Surgical Robots.
- Medical opportunities like adaptable pacemakers, diagnostics (colonoscopies - catheters - guiding motion in uncertain environments).

**4. Material Advances:**

- Development of multifunctional materials, 'smart materials', variable stiffness materials, and metamaterials.
- Improved / combined /interfaced elastic + viscoelastic materials.
- Rapid fabrication advances, 3D printing.
- Active materials that utilize energy reservoirs for adaptive behaviors.

**5. Sensing and Actuation:**

- Embedded sensors that enable general-purpose applications.
- More sophisticated tactile sensors with embodied characteristics.
- Improved actuators that can perform mechanical computations and adaptation as they actuate.

**6. Energy Efficiency and Sustainability:**

- Reduction of power consumption.
- More energy-efficient robots that require less compute.

**7. Security:**

- Development of secure robots that cannot be hacked.

**8. Other Applications:**

- Application in other domains beyond robots (e.g., intelligent buildings).
- Robots that react and eliminate pests in agriculture.

**Timeline:**

- 5-10 years: Subsystem / component level advancements.
- 5 years: Multiscale Hierarchical Integration of sensors, mechanisms, and controls.
- 10 years: Medical applications, exosuits, assistive robotics, embodied AI / industry involvement.
- 20 years: "True co-bots", ubiquitous deployment of robots, more direct interactions with the environment (e.g., drones perching), transistor-free robots.

These themes suggest a future where MI will play a significant role in various fields, from robotics and medicine to materials science and energy efficiency. The experts also highlight the importance of bio-inspired design and the integration of mechanical, computational, and possibly even biological intelligence. They foresee advancements at both the subsystem/component level and the system level, with a timeline spanning the next 5 to 20 years.

## 2. What are the impacts of these foreseen advances?

The impacts of the foreseen advances in Mechanical Intelligence (MI) can be summarized as follows:

**1. Robust and Adaptable Systems:**

- Systems that adapt to challenging environments and are robust.
- Robots and AI that aren't brittle or vulnerable to adversarial attacks.
- Long-term adaptable systems that 'grow' with time.
- Robustness of machines and personally adapted robotics.

**2. Performance and Efficiency:**

- New capabilities and higher performance.
- Lower/no power consumption, more reliable systems.
- More efficient use of power and energy.
- Providing hardware systems that are actually good for high-level computation/AI to be truly useful in complex, diverse scenarios.

**3. Cost and Accessibility:**

- Lower cost systems that increase access to technology.
- Cost production drops leading to lower cost robots.
- Accessibility improvements in technology.

**4. Biocompatibility and Sustainability:**

- Biocompatible systems in 20 years.
- Biointegrated prosthetics, exoskeletons, artificial endoskeleton.
- Sustainable systems in 10 years, including self-powered systems.
- More sustainable manufacturing and disposable/dissolvable robots.

**5. Real-world/Societal Impact:**

- Impact on healthcare, search and rescue, and education sectors.
- Expansion of the consumer robotics market and application of robots.
- Development of medical devices and space applications.
- Agriculture advancements like ultra-targeted pesticide delivery.

**6. Technology and Material Advances:**

- Technology at the component/material level - preintegration.

- Scalability and manufacturability improvements.
- Development of new materials, sensors, actuators.
- Digital-to-actual design flow advancements.

These impacts suggest a future where MI will lead to robust and adaptable systems, improved performance and efficiency, lower costs and increased accessibility, biocompatible and sustainable solutions, and significant real-world/societal impacts. The experts also highlight the advancements in technology and materials that will drive these impacts.

## 3. What basic research directions or questions can research in mechanically intelligent systems address?

The key research directions or questions that research in mechanically intelligent systems can address can be summarized as follows:

**1. Understanding and Defining Intelligence:**

- Answering "what is intelligence" (perhaps through biology).
- Is there a unifying theory of how to apply mechanical intelligence or even a unifying metric of evaluation?

**2. System Design and Integration:**

- Systems integration and real-time adaptation.
- Optimizing mechanical and computational effort.
- How much of a robotic task is preplanned or explicitly decided vs incorporated in the mechanism can be adjusted.
- How to co-design mechanical intelligent systems that can satisfy its functional requirements and intelligence/computational power?
- How to integrate mechanical and digital intelligence?
- How to use control systems in collaboration with mechanical intelligence?

**3. Material and Structural Advances:**

- An expanded understanding of the highly nonhomogeneous properties of biological tissues.
- Ability to 3D print heterogeneous materials and structures at fine scale.
- Develop new designs that exploit mechanical intelligence at the microscale.
- New multifunctional materials that can think - How can we turn materials into components to achieve mechanical intelligence?
- How do we learn the correct multimaterial and multiscale distributions to meet desired local and global behaviors?
- How to fabricate these complex multimaterial, multiscale structures cheaply, at scale?

**4. Control and Sensing:**

- Simplify control by providing default robustly stable behavior.
- Distributed sensing.
- Minimally invasive procedures will require increasingly small tools that are inherently safe and biocompatible while also accomplishing desired function.

**5. Computational Capabilities:**

- Computational capabilities using mechanical systems (e.g. solving PDEs).
- General "compilers" for building solutions/capabilities into mechanical structure.
- Modeling and simulation of contact, interfaces of different materials/multimaterials.
- Quantify the benefits of MI for better integration with AI.

**6. Application and Deployment:**

- Deploy system in new environments - interact with more environments.
- Mechanical Intelligent systems would influence basic research questions in Biology/Organismal biology/Behavioral biology, Tissue and organ engineering, Neuroscience/Artificial Intelligence/Embodied Intelligence, Robotics and robots that interact and learn autonomously, Materials and meta-materials, including novel sensors and actuators.

**7. Future Research Directions:**

- What are necessary characteristics for mechanical intelligence, what are the substrates to accomplish this?
- Mapping information theory into the mechanically response of the substrates to understand how is information flowing and how much information do we have.
- Quantifying material properties in terms of information processing.
- What does a system theory look like when there's no block diagrams?
- When you can't divide the world into pieces how do you glue things together (e.g., materials and design choices)?
- If traditional mechanical systems rely on energy, intelligence systems rely on entropy and we need a theory that encompasses both.
- Modeling of heterogeneous systems that are not precise/predictable but are repeatable.

These themes suggest a future where research in MI will address a wide range of topics, from understanding and defining intelligence, system design and integration, material and structural advances, control and sensing, computational capabilities, application and deployment, to future research directions.

# 4. What are the current barriers to advancing research in mechanical intelligence?

The current barriers to advancing research in Mechanical Intelligence (MI) can be summarized as follows:

**1. Complexity and Understanding:**

- Complexity of the properties of active materials (tissues, muscles, cilia, etc.).
- Hierarchical and coupled feedforward and feedback regulation mechanisms across length and time scales.
- Need for an information theory or an understanding of information transfer and communication for mechanical systems.
- Need for new theories for analyzing active mechanical systems in changing mechanical environments.

**2. Design and Methodology:**

- Development of a "world model" - a statistical understanding of how the robot or an "active" mechanical system will interact with the world.
- Formal design methods for mechanical/reflexive/high level control hierarchies (Co-design?).
- How to incorporate activity, geometry/morphology, material properties, sensory networks to develop intelligent physical systems?
- Need for tools for rapidly and systematically exploring large design (and fabrication?) space.

**3. Standards and Benchmarks:**

- Lack of established benchmarks.
- Need for standardization, prototype, physical system, that enables theoretical/model based advances to actually test on real systems in an accessible way.
- Need for standardized tasks – demos that are well accepted and where people who develop hardware can demonstrate performance.

**4. Funding and Acceptance:**

- Acceptance of the field - Is there a research centralization for these topics in academia? Where is MI research being presented and published and do researchers across disciplines "see" each others work?
- Funding challenges - Is there a funding "home" for MI that will support this work long term?

**5. Scalability and Manufacturability:**

- Scalability and manufacturability research to change lab concepts to more realistic component fabrication.
- Co-design methodology? Functional requirement and intelligence requirement to create an efficient system: What methodology are you going to use? What kind of optimization?

**6. Interdisciplinary Collaboration:**

- Hugely multi-disciplinary right now - brain + body + biology.
- The breadth of potential areas of application makes finding commonality between those areas to work together.
- Attitudes amongst some researchers are silo'ed especially well-funded areas - making the interaction with interdisciplinary researchers difficult.
- Cultural barriers (e.g. language) between disciplines (conferences could help?).

These barriers suggest that advancing research in MI requires addressing challenges in understanding complexity, improving design methodologies, establishing standards and benchmarks, securing funding and acceptance, enhancing scalability and manufacturability, and fostering interdisciplinary collaboration.

# 5. Discussion Session 3 Takeaway Summary

The key takeaways from the discussion on advancing research in Mechanical Intelligence (MI) during Session 3 can be summarized as follows:

**1. Material and System Integration:**

- Barriers for systems people to incorporate materials that are not packaged in actuators and sensors.
- The potential of Biology-Systems engineering research, despite cultural and funding-agency barriers.
- The need for multi-material 3D printing and bio-inspired fibrous heterogeneous structures to exploit material heterogeneity towards mechanical intelligence at the microscale.

**2. Optimization and Co-design:**

- The current emphasis on Machine Learning (ML) optimization may lead to underutilization of optimization for MI-based design.
- The importance of co-design, involving multiple optimizations (e.g., dynamics and controller design), as an integral feature of advances in MI.
- The value of MI comes from the context of the system, highlighting the importance of systemic issues in co-design.

**3. Personalization and Ubiquity:**

- MI has the potential to allow personalization and ubiquity of autonomous systems.

- Commercialization and the breadth of disciplines can make progress difficult.

**4. Benchmarking and Metrics:**

- There is a dire need for benchmarking, including benchmark tasks, benchmark metrics (including adaptability), and strong baselines of comparison to show the benefits of mechanical intelligence.
- The need for a systematic process for the "discovery"/"definition" of benchmarks.
- Measuring MI by equivalent Intelligent system (noting sensors/actuator replacement more than the number of lines).

**5. Societal Impact and Research Challenges:**

- Significant societal impact is expected from advances in MI.
- To achieve this, there are system-level and component/material-level research challenges, including modular architecture, manufacturability, scalability, and co-design methodology.
- The need for a hierarchical architecture involving preflex, reflex, and higher-level control.

**6. Materials and Control:**

- The potential of metamaterials to go beyond biomaterials, functioning as a whole system from a materials perspective.
- Mechanical intelligence should complement or ease control.

**7. Theoretical Underpinning and Integration:**

- The need for theoretical underpinning in Physics (active material), Chemistry, and co-design methodology.
- The importance of integrating actuation, sensing, power, and compute into single parts, requiring new design tools/a new systems theory for MI.

**8. Bio-compatibility and Sustainability:**

- The importance of bio-compatibility and sustainability in MI research.
- The need for integrative hierarchical sensing and control, with mechanical, reflexive, and cognitive aspects.

These takeaways highlight the importance of material and system integration, optimization and co-design, personalization, benchmarking, addressing societal impact and research challenges, understanding materials and control, theoretical underpinning, and focusing on bio-compatibility and sustainability in advancing research in MI.

# Keynote summaries

# Overview of Keynote 1 and 2

The first keynote was given by Dr. Barry Trimmer. The presentation discussed the definition and diversity of intelligence, the relationship between mechanical intelligence and morphological computation, the role of materials, structure, and control in intelligent systems, and identified areas for further exploration, including time-dependent changes and autonomous problem-solving.

The second keynote was given by Dr. Hannah Stuart. The presentation discussed the role of risk, environment, and embodied reactions in mechanical intelligence, the limitations of non-mechanical intelligence, the need for complex measurements in evaluating embodied intelligence, and posed questions about the measurement, observation, and manipulation of mechanical intelligence.

## Q&A summary

Following the first two keynotes an open discussion and questions session was held. A summary of the keypoints of this discussion follows here:

1. **Definition and Scope of Mechanical Intelligence**: The discussion started with a query about the limits of mechanical intelligence, particularly in relation to the nonlinearity of elasticity in biological systems. It was suggested that any mechanical system that can acquire knowledge, process it, and respond purposely can be defined as having mechanical intelligence. The conversation then focused on the scope of mechanical intelligence, arguing against limiting it to just locomotion or manipulation. It was suggested that mechanical intelligence encompasses a spectrum of smaller and bigger circles.
2. **Design Space and Task Specificity**: The conversation moved to the challenge of defining the design space in this field, considering factors like task specificity, scale lengths, time, and the complexity of structures. The idea of creating a design space to identify potential advances was proposed.
3. **Role of Environment and Energy Harvesting**: The environment was recognized as a crucial part of the parameter space or design space. Organisms, having evolved in an environment, are inseparable from it. The environment plays a key role in aspects like energy harvesting and interaction with physical variables. Coming from an energy harvesting background, one participant emphasized the importance of empowering the system and how biological systems interact with the environment and external stimuli for energy harvesting.
4. **Advantages of Mechanical Compliance and AI**: It was pointed out that despite the slow movement of caterpillars, mechanical compliance offers immediate response, which can be extremely fast. This was exemplified by a cockroach compensating for a sudden force faster than its nervous system could react. The discussion explored the potential of mechanical intelligence and AI complementing each other, particularly in the context of soft robotics. The question was raised about how controls could be leveraged to improve aspects of soft robots, which are typically slow due to their mechanical compliance.
5. **Role of Transducers and Sensing in Mechanical Intelligence**: The role of transducers in mechanical intelligence was discussed. It was clarified that transducers are not limited to transforming mechanical signals into electrical signals, but can also involve transformations between other types of signals, such as pressure to force. This led to a debate about whether transducers without electrical signals involved would still fall within the definition of mechanical intelligence. The conversation then focused on the role of sensing in mechanical intelligence. It was argued that separating sensing from mechanical intelligence goes against the principle that sensing should be integrated within mechanical intelligence. Examples from biology were given, such as the basilar membrane in our ears, where the mechanics of the sensors simplify control processing.
6. **Collective Intelligence**: The concept of collective intelligence was introduced, suggesting that all intelligence is a combination of many things and that trying to attribute intelligence to a single thing is futile. This led to the idea that all intelligence is ultimately collective, involving a multitude of interactions at various scales.
7. **Mechanical vs. Electrical Intelligence and Morphological Computation**: The discussion returned to the definition of a robot as an engineering construct that makes decisions based on measurements of the environment. This led to a debate about the threshold between mechanical and electrical intelligence, and whether computers could be considered intelligent. The conversation concluded with a discussion about the threshold between mechanical and electrical intelligence. It was suggested that the electrical part is what leads to computers, raising the question of whether computers are intelligent. The discussion ended with the idea that the definition of intelligence depends on how it is applied to a particular field. The conversation concluded with a debate about the distinction between mechanical intelligence and morphological computation. It was suggested that while a four-bar linkage or the physical structures in an owl's ear could perform a computation, they might not necessarily embody intelligence. The adaptability, responsiveness, and contextual dependency were seen as embodying intelligence. It was argued that morphological computation is part of mechanical intelligence, and that a mechanical system that can achieve the same goal through many different paths could be considered mechanically intelligent.

# Overview of Keynote 3 and 4

The third keynote was given by Dr. Sarah Bergbreiter. The presentation explored the expansion of the design space in mechanical intelligence, the challenges in navigating this space, the role of sensor mechanics and dynamics, and posed key questions about co-design, exploration of the design space, and the potential of mechanical systems to learn and surpass biological capabilities.

The fourth keynote was given by Dr. Todd Murphey. The presentation discussed the long-term goal of creating a transistor-less robot, the role of mechanical computing and smarticles control, the challenges and advantages of mechanical intelligence, potential metrics for evaluation, the concept of maximum diffusion reinforcement learning, and the need for a systems theory for mechanical intelligence as a design property.

## Q&A summary

Following the first two keynotes an open discussion and questions session was held. The discussion revolved around several recurring themes:

1. **Learning and Adaptability**: The discussion frequently returned to the idea of learning, both in terms of how mechanical systems learn their own dynamics and how they can adapt to new situations. This includes the concept of learning from diffusion and then moving to targeted learning.
2. **Design and Control**: There was a strong focus on the design aspects of mechanical intelligence, including the design of control actions, the design space, and the relationship between the design properties of control and the body.
3. **Degrees of Freedom (DOFs)**: The concept of DOFs was a recurring theme, with discussions around the number of DOFs required for mechanical intelligence, the inclusion of compute DOFs, and the potential use of DOFs as a metric.
4. **Human Involvement**: The role of humans in mechanical intelligence was also a key theme, with discussions around human-designed behavior, the value of human involvement, and the impact of human data.
5. **Metrics and Evaluation**: The discussion touched on various potential metrics for evaluating mechanical intelligence, such as control effort, learning efficiency, and ergodicity.
6. **Role of Environment and Physical Structure**: The influence of the environment and the physical structure of the system on mechanical intelligence was another important theme.

These themes reflect the complexity and multidisciplinary nature of mechanical intelligence, touching on aspects of learning, design, control, human involvement, and evaluation. They highlight the ongoing challenges and opportunities in this field.

More specifically, several ideas, metrics, and bottlenecks were introduced and debated:

- The statistical implications of diffusion were discussed, and it was noted that these match those by the controller.
- The importance of learning motion in shape space and locomotion in physical space was highlighted.
- The need to synthesize control actions to explore space was discussed, with the complexity of an agent learning its own dynamics being a key challenge.
- The idea of a learning hierarchy, starting from diffusion and then moving to targeted learning, was proposed.

- The spectrum from purely learned behavior to human-designed behavior was discussed, with the unpredictability of the evolution of purely learned behavior and the usefulness of human-designed behavior in situations with limited compute being noted.
- Questions were raised about when having a model matters, whether mechanical intelligence requires large numbers of degrees of freedom (DOFs), and the value of human involvement in mechanical intelligence.
- The potential for mechanical intelligence to make learning easier and better, to be used for input shaping on the sensor side, and to enhance control and processing capabilities was discussed.
- The importance of repeatability over predictability in mechanically intelligent systems was emphasized.
- The need for mechanical intelligence at small scales and the use of sensor-motor loops to take advantage of physical structure were discussed.
- Possible metrics for evaluating mechanical intelligence, such as control effort and DOFs, were proposed.
- The idea of a conserved landscape between design properties of control and the body was introduced.
- It was suggested that there will not be one definition of mechanical intelligence, but rather dimensions of mechanical intelligence.

# Report out Summary

During the final report out session, each small group was asked to provide a discussion synopsis, identify areas or opportunity, and identify areas of convergence with other fields. A summary of the reports outs from all groups is provided here.

## 1. Discussion Synopsis Summary

The overall discussion synopsis can be summarized as follows:

**1. Understanding and Defining MI:**

- MI lies on a spectrum with multiple axes, similar to other notions of intelligence.
- It provides responsiveness, robustness, efficiency, adaptability to challenging and changing environments.
- Mechanically intelligent systems are often characterized by a hierarchical structural/material organization.
- There is a lack of a fundamental theory for understanding, describing, and synthesizing these systems and their interaction with the environment.

**2. Benchmarking and Metrics:**

- Rather than a specific, common benchmark task and metrics, a range of benchmarks is needed.

- Quantifying MI is hard since intelligence can't be measured without the context (task/environment) that it's in.
- There is a need for measures that are more general than the performance of the task.
- There is a dire need for benchmarking, including benchmark tasks, benchmark metrics (including adaptability), and strong baselines of comparison to show the potential benefits of MI.

**3. Design and Methodology:**

- The challenge is to combine the components being developed now in an integrated system that has better performance.
- MI devices need to fulfill goals, through a diversity of behaviors.
- A constellation of metrics, specific to MI, will be part of the design process.
- There is a need for a systems theory and design processes that include both mechanics and information theory.

**4. Areas of Agreement and Disagreement:**

- There was agreement that MI is a mechanical system that has forces and motion performing a function that has aspects of intelligence and inherently involves the environment.
- There was disagreement on what level of intelligence should be considered MI and how to do benchmarking.
- There was no consensus on a single MI definition. However, there was broad consensus on the idea of mechanical intelligence as a continuum or spectrum. The idea being that relative to something else a system may be more or less mechanically intelligent. Furthermore, you may have sub-components that are mechanically intelligent (e.g. Mechanically intelligent sensing, mechanically intelligent grasping) even if as a whole the system relies on overarching embodied computational intelligence.

**5. Future Directions and Challenges:**

- MI will enable a new generation of autonomous systems that will enhance cyber security, direct environmental interactions, and energy efficiency.
- There are exploratory research opportunities across experiments, simulations, and computation.
- There are barriers for systems people to incorporate materials that are not packaged in actuators and sensors.
- There is a need for research focused on the integration part and for training and educational opportunities.

These takeaways highlight the importance of understanding and defining MI, the need for benchmarking and metrics, the challenges in design and methodology, the areas of agreement and disagreement among the participants, and the future directions and challenges in advancing research in MI.

## 2. Areas of Opportunity

The areas of opportunity identified by the workshop participants for advancing research in Mechanical Intelligence (MI) can be summarized as follows:

**1. Material Heterogeneity:**

- Exploiting material heterogeneity, enabled by advances in additive manufacturing, towards mechanical intelligence at the microscale.

**2. Mechanical Behavior and Control:**

- Exploiting potentially non-linear mechanical behavior to simplify distributed control and learning problems.
- Designing minimally invasive surgical tools that use intelligent mechanisms to enhance navigational autonomy while ensuring that fragile tissue will not be damaged.

**3. Benchmarking and Reproducible Research:**

- Early stage community buy-in for broadly defined benchmarking.
- Encouraging reproducible research where ad hoc benchmarking to other published works is both feasible and encouraged.

**4. Robot Agility and Task Efficiency:**

- Improved robot agility and ability to operate in challenging and extreme environments or do complex tasks.
- Solutions tailored to specific tasks that are robust and efficient.

**5. Beyond Robotics:**

- Adding intelligence to systems beyond robotics, such as adaptive/robotic buildings or clothing, and self-healing robots and structures.

**6. Sustainability and Life Cycle:**

- Improving the sustainability and life cycle of existing intelligent systems.
- Biodegradable and biohybrid robotic systems for improved sustainability and adaptability

**7. Design Tools and Co-design:**

- Multidisciplinary design tools for integrated MI systems.
- Exploration of new properties for materials and mechanisms that can be used for MI systems.
- Co-design of mechanical properties, functionality, and compute at the materials and component level, and at the system level.

**8. Scalability and Manufacturability:**

- Scalability and manufacturability research to advance lab concepts to more realistic components and systems.

**9. Sensor Integration and Control:**

- Multiscale Hierarchical Integration of sensors, mechanisms, and controls.
- More sophisticated tactile sensors with embodied characteristics.

**10. Medical Applications and Biocompatibility:**

- Medical applications, exosuits, assistive robotics, and "True co-bots".
- How to design biocompatible machines/implantable devices, this could include biohybrid robotics.

**11. Theory Development:**

- A mechanical theory of general computation to enable co-design.
- Developing a novel theory of mechanical intelligence that draws upon tools from information theory, active matter physics, artificial intelligence and machine learning, classical mechanics and metamaterials, and control theory and dynamical systems.

These areas of opportunity suggest a future where research in MI will address a wide range of topics, from material heterogeneity, mechanical behavior and control, benchmarking, robot agility and task efficiency, sustainability, design tools and co-design, sensor integration and control, medical applications, and biocompatibility, to theory development.

## 3. Convergence with other fields

Participants overall felt that Mechanical intelligence is an inherently multi-disciplinary research area requiring collaboration across many fields (mechanics, mechanisms, dynamics, controls, materials science, robotics, learning, computing, manufacturing, chemistry, materials science, biology, math, computing, etc).

Based on the overall report outs, the fields that Mechanical Intelligence (MI) is most likely to converge with can be summarized as follows:

**1. Material Science and Manufacturing:**

- Multi-material 3D printing.
- Bio-composite, bio-hybrid inks.
- Multi-functional inks.
- Basic science for new materials, bio-inspired principles/materials.

**2. Biology and Biotechnology:**

- Harnessing biological self-assembly.
- Mechanobiology.
- Leverage increasing understanding of engineering living materials to create life-like and living robots.

**3. Control Theory and Robotics:**

- Distributed and morphological control.
- Systematic approaches for integrating MI and AI.
- Think about MI outside of robotics (e.g., smart packaging, agriculture, medicine, space, civil infrastructure).

**4. Computational Modeling and Optimization:**

- Computational modeling of heterogeneous materials.
- Multi-objective design optimization where the optimization problem is framed to meet critical metrics (e.g., robustness).
- Leverage AI/computational tools to explore large multi-design spaces.

**5. Environmental Engineering and Sustainability Studies:**

- Applications related to environmental engineering/sustainability studies.
- Completely recyclable robots.

**6. Human Factors and Social Sciences:**

- Human factors studies needed for developing applications.
- Social science researchers to understand impact to society, ethics of these systems.

**7. Security and Safety:**

- "Unhackable" mechanical systems.

These convergence areas suggest a future where research in MI will intersect with a wide range of fields, from material science and manufacturing, biology and biotechnology, control theory and robotics, to computational modeling and optimization, multidisciplinary approaches, environmental engineering and sustainability studies, human factors and social sciences, specific industries and professions, and security and safety.

# Acknowledgments

The organizers would like to thank all workshop participants for their lively and engaging discussion. The ideas reflected here are directly summarized from their discussions. Microsoft CoPilot with commercial data protections through Carnegie Mellon University was used to cluster participant responses in order to efficiently find themes that emerged across groups.

# Epilogue

At the end of the workshop, two participants (Zeynep Temel and Margaret Coad) came across a street poet in Alexandria and commissioned a poem on mechanical intelligence. We end our report with this poem below.

**"Mechanical intelligence"**

the devices we made, they understand
How patterns wait
behind this universe

a spiderweb we glimpse

    as symbols
        as sounds

this mathematics

    a fiber sturdy enough

        and so fundamental

    You could build a bridge
        out of it

    across the dreary old gap
    between what is

        and what could be.

Author,
Tristan Varna